\def\year{2018}\relax
\documentclass[letterpaper]{article} 
\usepackage{aaai18}  
\usepackage{times}  
\usepackage{helvet}  
\usepackage{courier}  
\usepackage{url}  
\usepackage{graphicx}  
\newcommand{\eg}{\textit{e.g.}}

\newcommand{\etc}{\textit{etc.}}
\newcommand{\ie}{\textit{i.e.}}

\usepackage{subcaption}
\usepackage{epsfig}
\usepackage{amsmath}
\usepackage{amssymb}
\usepackage{amsthm}
\usepackage[table]{xcolor}

\usepackage{bm}
\usepackage[mathscr]{eucal}

\begin{document}

\title{Brute-Force Facial Landmark Analysis With A 140,000-Way Classifier}

 \author{Mengtian Li \qquad Laszlo Jeni \qquad Deva Ramanan\\
 The Robotics Institute, Carnegie Mellon University\\
 {\tt\small \{mtli, laszlojeni, deva\}@cmu.edu}
 }

\maketitle

\begin{abstract}
We propose a simple approach to visual alignment, focusing on the illustrative task of facial landmark estimation. While most prior work treats this as a regression problem, we instead formulate it as a discrete $K$-way classification task, where a classifier is trained to return one of $K$ discrete alignments. One crucial benefit of a classifier is the ability to report back a (softmax) distribution over putative alignments. We demonstrate that this distribution is a rich representation that can be marginalized (to generate uncertainty estimates over groups of landmarks) and conditioned on (to incorporate top-down context, provided by temporal constraints in a video stream or an interactive human user). Such capabilities are difficult to integrate into classic regression-based approaches. We study performance as a function of the number of classes $K$, including the extreme ``exemplar class'' setting where $K$ is equal to the number of training examples (140K in our setting). Perhaps surprisingly, we show that classifiers can still be learned in this setting.
When compared to prior work in classification, our $K$ is unprecedentedly large, including many ``fine-grained'' classes that are very similar. We address these issues by using a multi-label loss function that allows for training examples to be non-uniformly shared across discrete classes.
 We perform a comprehensive experimental analysis of our method on standard benchmarks, demonstrating state-of-the-art results for facial alignment in videos.
\end{abstract}

\section{Introduction}
\setcounter{secnumdepth}{2}

Accurately localizing facial landmarks is a core competency for many applications such as face recognition, facial expression analysis and human-computer interaction. 
Performance of existing methods is quite impressive on datasets captured in constrained scenarios. As such, attention in the community has shifted towards ``in the wild'' \cite{Sagonas2016300FI} 
settings, for which large pose variation and severe occlusions pose significant challenges. While numerous attempts have been made to address them,
our evaluation suggests that these harder problems are far from being solved.

\begin{figure}[t]
\centering
\includegraphics[width=1\linewidth]{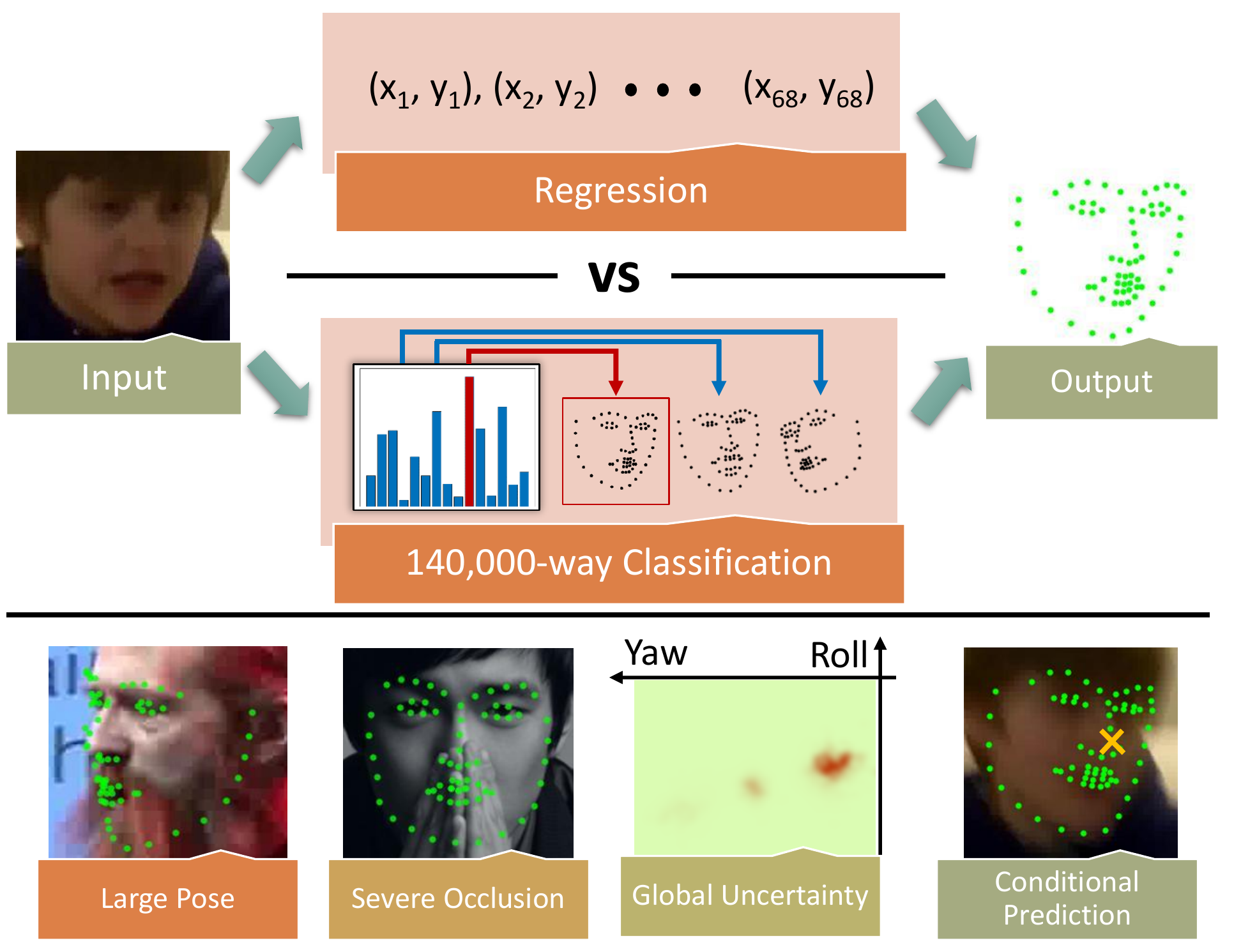}
\caption{Face alignment is a regression problem, yet we solve it via large-scale classification.  As shown in the bottom row, our model is able to handle severe occlusions and large pose variation and provide a global uncertainty estimate. Moreover, such uncertainty representation can used to produce conditional prediction in an interactive setup.}
\label{fig:teaser}
\end{figure}

{\bf Motivation:} To address these remaining challenges, let's take a step back. Computer vision can be thought of as an inverse estimation problem, where given an image, one has to estimate a high-dimensional set of parameters specifying the true underlying geometric properties of the scene (in our case, facial landmarks). Such inverse estimation problems are notoriously difficult, but are currently enjoying a period of transformative success due to data-driven architectures such as deep networks. While enormously powerful, such networks reduce the problem to one of nonlinear regression. Such an approach may suffer when the inverse problem is inherently {\em ill-conditioned}, implying that multiple interpretations/solutions may be equally valid. In this case, it may be more natural to predict a {\em distribution} over interpretations. For example, when applying such networks to predict the location of landmarks in a heavily occluded face, 
it likely helps to report back multiple possibilities.

{\bf Approach:} Given the above motivation, we propose a simple but somewhat radical approach to alignment: discretize {\em all} possible predictions into $K$ discrete classes, and treat the problem as one of large-scale $K$-way classification. Since networks are readily trained to report back (softmax) distributions over classes, this approach produces uncertainty estimates over possible interpretations of an image. Importantly, such an approach also requires scaling up classification networks to massive number of classes (in the hundreds of thousands or millions), which poses several theoretical and practical challenges. 

{\bf Scalability:} To the best of our knowledge, no work has attempted to solve a classification problem at this scale for visual understanding problems. The closest work is \cite{Dean2013FastAD} and \cite{joulin2016learning} \footnote{A large set of classes also appears in face recognition literature. However the problem is usually formulated as identification and verification \cite{KemelmacherShlizerman2016TheMB}, different from direct classification.}.
The first work trains deformable-part models to detect 100,000 object classes. The second trains a network with 100,000 classes in the unsupervised fashion using a loss that is equivalent to a stochastic version of our soft target, which is superseded by multi-label loss in our work. \cite{Dean2013FastAD} uses MapReduce framework, \cite{joulin2016learning} uses 4 GPUs, while our work uses a single GPU in MATLAB and handles 40\% more classes. In our work, we show that deep networks partially address the challenge of computation
by a remarkable ability to share computation across multiple tasks (or classes). Our experiment shows that at test time, the 140K-class setting has negligible increase in forward pass time compared to the 10-class one, since the bulk of the computation is feature extraction. Another challenge is performance -- it is difficult to learn decision boundaries across ``fine-grained" classes that are very similar. To address this challenge, we introduce a {\em multi-label} framework for training {\em multi-class} networks at scale. Importantly, our framework allows for training examples to be non-uniformly shared across classification tasks. 

{\bf Capturing uncertainty:} In contrast to regression-based methods for alignment, our approach has the unique ability to report back {\em joint distributions} over landmarks. This allows for a variety of novel operations. Firstly, our system can report back uncertainty estimates in global variables of interest, such as viewpoint. Secondly, our system can report back {\em conditional} distributions by conditioning on knowledge provided from top-down context. We focus on alignment in video sequences, where temporal context can be used to refine uncertainty estimates in an individual frame (that may be ambiguous due an occlusion). We also show that humans can provide such top-context, allowing our system to be used as an interactive annotation interface. With a single user-click, our system produces near-perfect landmark accuracy.

{\bf Evaluation:} We evaluate our $K$-way classification network for the task of facial alignment in video sequences, focusing on the recent 300 Videos in the Wild (300VW) benchmark \cite{shen2015first,Chrysos2017ACP}. To explore the impact of large K, we make use of 140,000-image training set consisting of real images and publicly-available synthetic images obtained by pose-warping the real training set~\cite{Zhu2016FaceAA}.
We demonstrate state-of-the-art accuracy in terms of coarse alignment, as measured by the number of frames where landmarks are localized within a coarse tolerance. This is somewhat expected as our outputs are discretized by design. To improve accuracy for small tolerances, we add a post-processing regression step that produces state-of-the-art results across all tolerance thresholds.



\section{Related Work}
\setcounter{secnumdepth}{2}

Automatic face alignment has been an active area of computer vision. During the last few decades the field underwent major changes both in methodology and in operating conditions. Early works can usually be categorized into Active Shape Models (ASM) \cite{Cootes1992,Cootes1993}, Active Appearance Models (AAM) \cite{Cootes1998,Gross2005,Matthews2004} and Constrained Local Models (CLM) \cite{Saragih2011,Sangineto13,Baltrusaitis2012,Yu2013}. The emergence of Cascaded Regression Methods (CRM) ~\cite{Cao2013,Yang2013,Xiong2013,Xiong2015,Tzimiropoulos2015,Zhu2015,Yang2015FacialST,Deng2016M3CM} brought significant performance gain in fitting speed and accuracy ~\cite{Kazemi2014,Ren2014}. In recent year, deep learning based methods further improved precision and robustness on challenging cases. \cite{Fan2016ApproachingHL} employed multiple CNNs in a coarse-to-fine fashin. \cite{Zhang2016SP} adopted multi-task learning in their cascaded CNN framework. \cite{Zhu2016FaceAA} treated $(x, y, z)$ coordinates as RGB values and along with the image, fed it into a CNN, which iteratively refines the underlying facial parameters. Other work incorporated recurrent models \cite{Trigeorgis_2016_CVPR,Peng2016ARE} or generative models \cite{Zhang_2016_CVPR}.

How these various methods compare in more challenging real-life video scenarios was relatively unknown. There was not a commonly accepted evaluation protocol or enough annotated data for joint face tracking and alignment until the release of the 300VW benchmark \cite{shen2015first,Chrysos2017ACP}. This benchmark contains more than 100 annotated videos and aims to evaluate facial landmark tracking in both constrained and unconstrained settings.

Several methods have been proposed to address this challenging task.  \cite{Yang2015FacialST} employed a spatio-temporal cascaded shape regression that combined multi-view regression with time-series regression to improve temporal consistency. \cite{uricar2015real} used a Deformable Part Model detector extended with a Kalman filter for temporal smoothing. \cite{Xiao2015FacialLD} presented a multi-stage regression-based approach, that progressively initializes the more challenging contour features from stable fiducial landmarks. \cite{Rajamanoharan2015MultiviewCL} proposed a multi-view CLM that employs a global shape model with head-pose specific response maps. \cite{Wu2015ShapeAR} proposed an approach to better utilize the shape information in cascade regressors, by explicitly combining shape with appearance information.  In the online setting, \cite{SnchezLozano2016CascadedCR} proposed a cascaded continuous regression that can be updated incrementally.

\section{Regression by Large-Scale Classification}
\setcounter{secnumdepth}{2}

Given an image with a roughly-detected face ${\bf I}$, we wish to infer a set of $N$ landmark points. Instead of treating this as a continuous regression problem,
\begin{align}
    f({\bf I}) = {\bf y}, \quad {\bf y} \in {\bf P} = \mathbb{R}^{N \times 2} \quad \text{[Regression]}
\end{align}
we convert it into a $K$-way classification problem:
\begin{align}
    f({\bf I}) \in \{ {\bm \mu}_1,\ldots, {\bm \mu}_K \}, \quad {\bm \mu}_k \in {\bf P} \quad \text{[Classification]}
\end{align}
Intuitively, pose classes may capture the variation of faces along {\em pose, expression, and identity}. We note that the above formulation can actually be relaxed into arbitrary annotations for each discrete class. For example, different pose classes may contain different number of visible points, implying that the reported landmarks $\mu_k$ need not lie in the same space ${\bf P}$. Nonetheless, we will assume this for notational simplicity.

\label{sec:poseclass}

\begin{figure}[t!]
\begin{center}
   \includegraphics[width=1\linewidth]{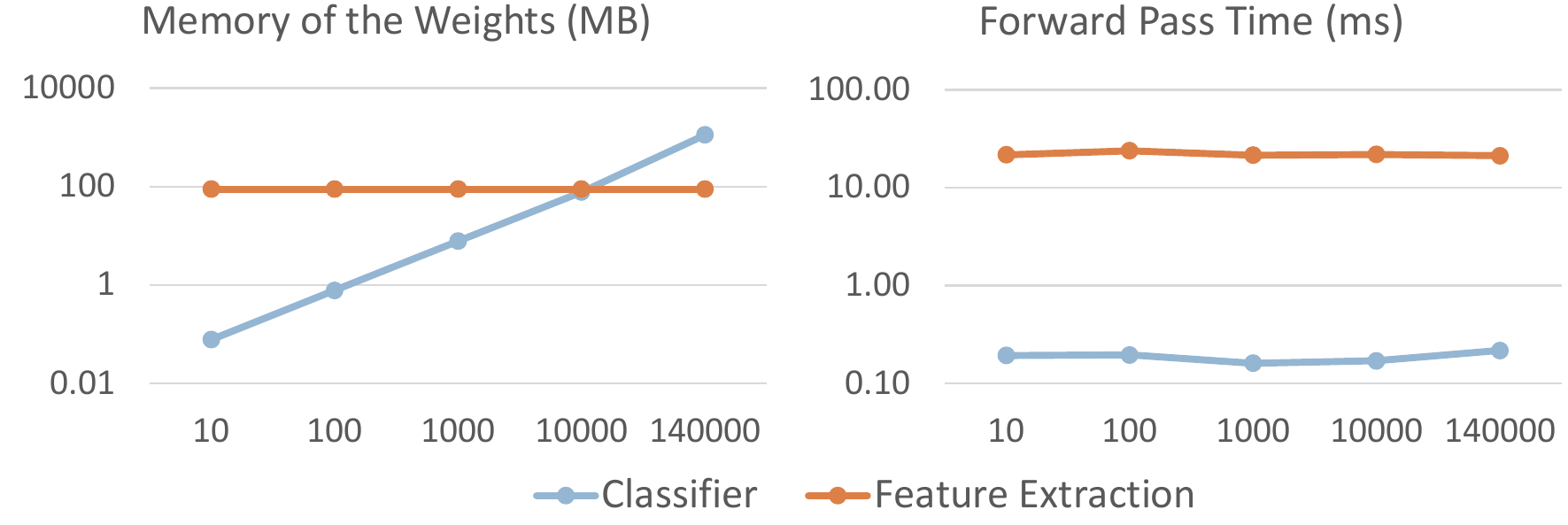}
\end{center}
   \caption{The computational constraints of our model at test time. Orange represents all layers prior to the last fully-connected layer (feature extraction), and blue represents the last fully-connected layer (classifier). With increasing number of classes, the running time barely increases while the memory consumption increases linearly.}
\label{fig:computation}
\end{figure}

{\bf Clustering:} Given a training set of $M$ face images and associated landmark annotations $({\bf I}_i, {\bf y}_i)$, we first must generate a set of $K$ discrete pose classes. To do so, we center and scale all landmarks by aligning the ground truth detection window and perform $k$-means clustering to generate the set $\{{\bm \mu}_k\}$.
We consider various values of $K$, including $K=M$, corresponding to singleton clusters (in which no clustering need actually happen).

{\bf Probabilistic reasoning:} We will explore classification architectures that return (softmax) probability distributions $p_k$ over $K$ output classes. We convert this to a distribution over landmarks with a mixture model:
\begin{align}
\label{eq:joint}
   p({\bf y}) \propto \sum_k  p_k \phi_k(||{\bf y}-{\bm \mu}_k||),
\end{align}
where $\phi_k$ is a standard {\em radial basis function}. We use a spherical Gaussian kernel fit to the $k^{th}$ cluster.  The joint distribution allows us to perform standard probabilistic operations such as {\em marginalization} and {\em conditioning}.  We can compute marginal uncertainties over individual landmarks (\eg, ``heat maps'') or groups of them (\eg, uncertainty estimates over global properties such as viewpoint - see Fig \ref{fig:teaser}). We can condition on evidence provided external constraints (arising from temporal context or interactive user input), as shown in Sec~\ref{sec:interactive}. 


To derive our final scalable approach, we will first describe ``obvious" strategies and analyze where they fail, building up to our final solution.

\begin{figure}[t!]
\begin{center}
   \includegraphics[width=1\linewidth]{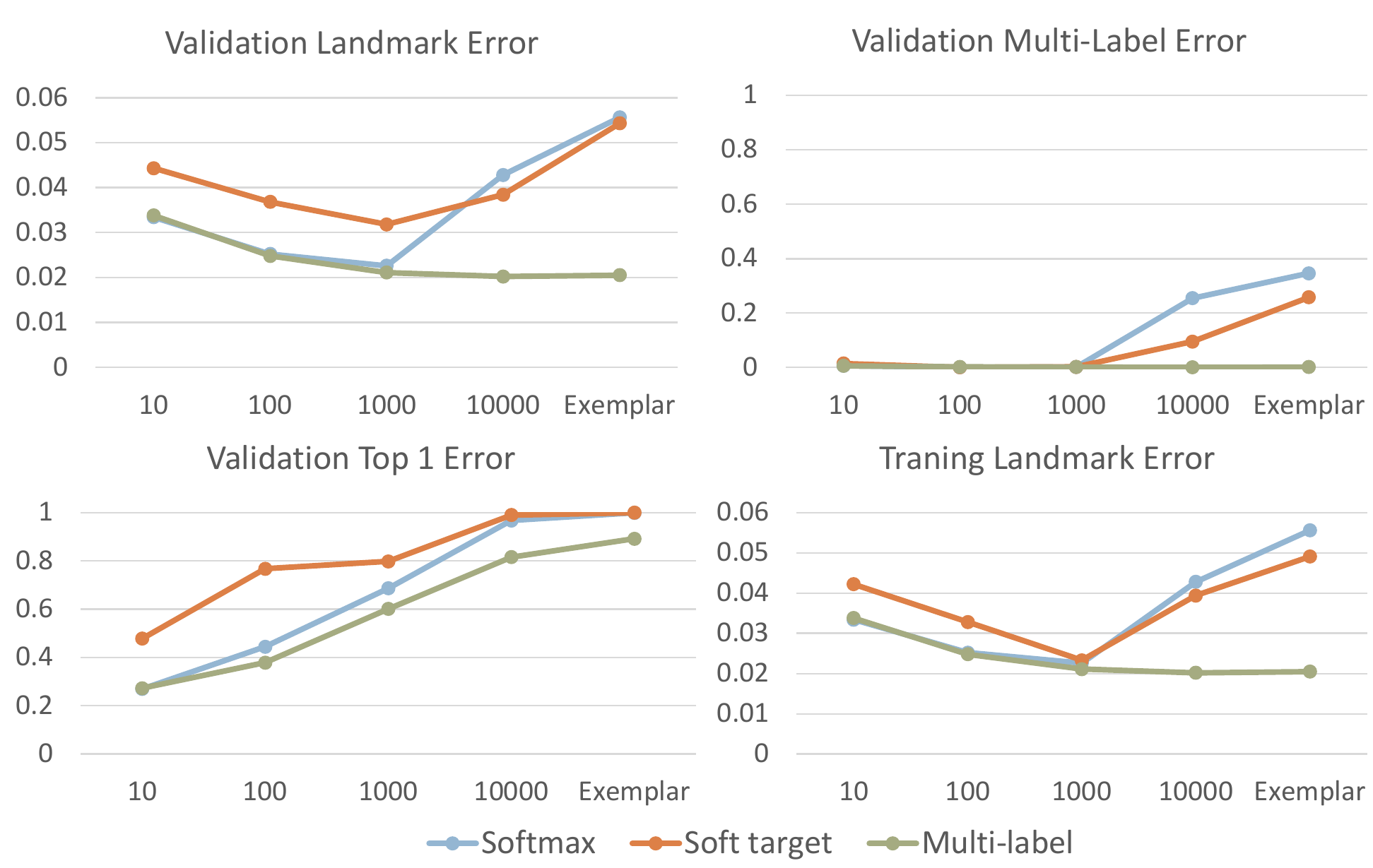}
\end{center}
   \caption{Effect of different training losses with increasing number of classes. The landmark error is the standard normalized pt-pt error (RMSE). For multi-label error, we count the prediction incorrect only if it is not a member\ of the pose class. We see that the commonly used softmax log loss and the soft target does not scale up with the number of classes. We adopt multi-label loss for our classification network. In this diagnostic experiment, only real images are used during training and the number of exemplars is around 26,000.}
\label{fig:increasingK}
\end{figure}

\subsection{Attempt 1: Naive $K$-way Classification}

With our discrete classes defined, we are ready to train a $K$-way classifier! We begin by training a standard deep classification network (ResNet~\cite{He2016DeepRL}) for $K$-way classification on pose classes. Because it will prove useful later, let us formalize the standard cross-entropy loss function commonly used to train $K$-way classifier. Let $s_k({\bf I})$ be the prediction score of class $k$ for image {\bf I},  
 ${\bf p}$ be the predicted distribution and ${\bf q}$ be the target distribution, then the {\em cross-entropy loss} is 
\begin{align}
\label{eq:crossentropy}
H({\bf p}, {\bf q}) = -\sum_{k = 1}^{K} q_k \log p_k, \ p_k = \frac{\exp(s_k({\bf I}))}{\sum_{j = 1}^{K}{\exp(s_j({\bf I}))}}.
\end{align}

Typically, the target distribution is a one-hot-encoded vector specifying the pose class of this training example:
\begin{align}
\label{eq:softmax}
\text{SoftmaxLoss} = H({\bf p}, {\bf q}), \ q_k =\delta(k = c)
\end{align}
with $c$ being the ground truth class.

{\bf Computation:} Perhaps our first surprising conclusion is that such architectures {\em do} scale to such massive $K$ (140,000), at least from a computational point-of-view. One would imagine that the classification time would increase given the large number of classes we have. As shown in Fig \ref{fig:computation}, however, a larger number of classes has negligible increase in the running time but consumes much more memory. This might be a result of modern GPUs being well-optimized for convolutions. Therefore, the scalability of having more classes is mainly constrained by the 12GB memory available on current graphics cards.

{\bf Performance:} Performance is plotted as a function of $K$ as the blue curves in Fig \ref{fig:increasingK}. One immediate observation from the first column is that while normalized landmark error improves for some larger values of $K$, classification accuracy gets worse. The latter is not surprising in retrospect; a 100,000-way problem is harder than a $10$-way classification problem! Hence the appropriate evaluation measure seems to be landmark reprojection error. However, even when evaluating landmark reprojection error, performance maxes out at $K=1000$, but then drops dramatically, performing even worse than a 10-way model. Perhaps this also is not surprising in retrospect. As we increase the number of pose classes, we fragment the data more, to the point where each class contains a single example. Interestingly, fragmentation hurts not because of overfitting but because it makes the optimization problem more challenging (as evidenced by the increase in the training error in the bottom-right of Fig \ref{fig:increasingK}).

\subsection{Attempt 2: Soft Targets}
\label{sec:softtarget}

\begin{figure}[t]
\begin{center}
   \includegraphics[width=1\linewidth]{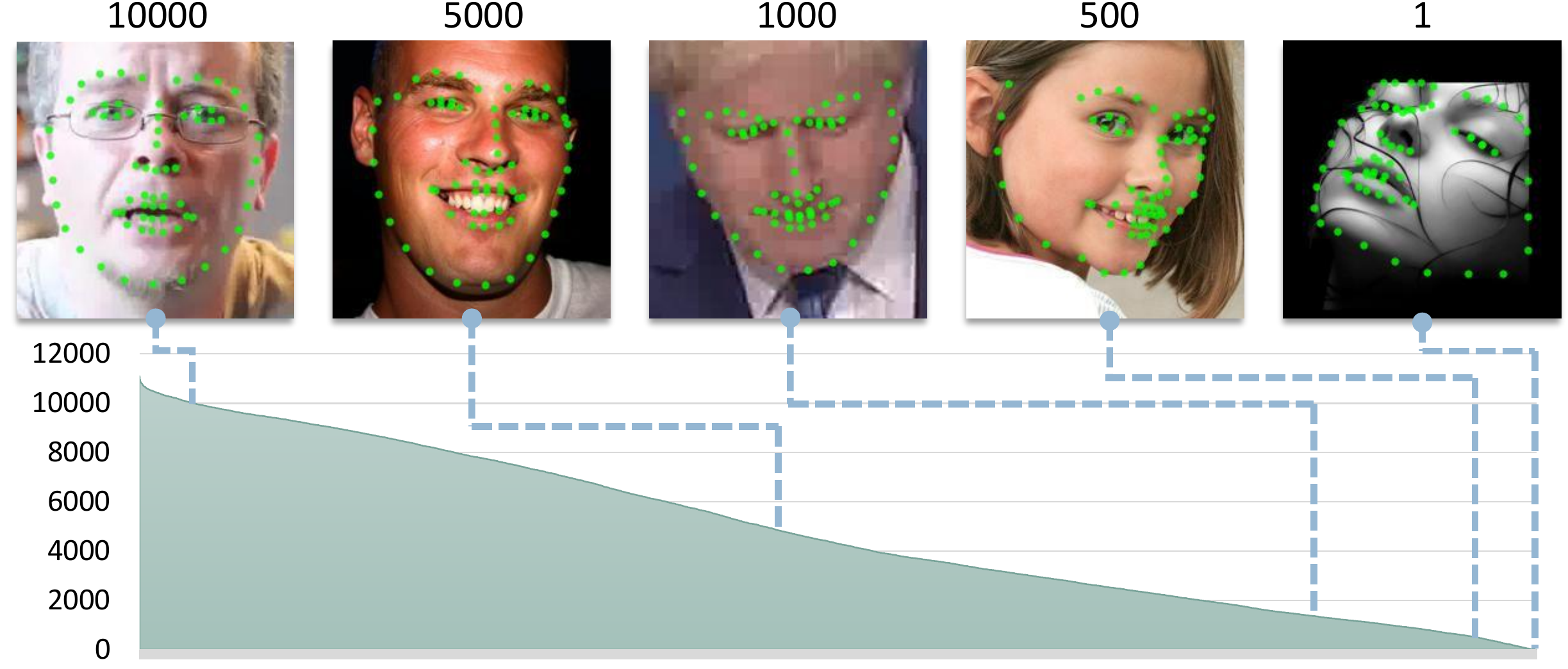}
\end{center}
   \caption{A distribution of images with different membership set sizes $|M_i|$. The positive set size represents the similarity with other examples and also the influence of the corresponding example at training time when using the multi-label loss.}
\label{fig:imgnbcnt}
\end{figure}

A related but subtly different hypothesis as to the poor scalability of large $K$ could be that the task simply becomes {\em too hard}.
Softmax requires that the exact correct label be returned, and if not, all other predictions are penalized {\em equally}. Instead, we may wish to train with some form of ``partial credit'' for reasonable predictions. To define the set of reasonable predictions for a training image $({\bf I}_i,{\bf y}_i)$, we find the set of pose classes that fall within some distance of ${\bf y}_i$, to form the {\em membership set}:
\begin{align}
    M_i = \{k: ||{\bm \mu}_k - {\bf y}_i|| \leq \tau\} \label{eq:members}
\end{align}
Conceptually, we can think of this as a ``growing" of the k-mean clusters to include shared examples (boxes in Row 3, Fig \ref{fig:classify}). Fig \ref{fig:imgnbcnt} ranks training examples by $|M_i|$. We see those with large memberships tend to be frontal faces with neutral expressions, while those with few examples tend to be extreme poses.
One approach for partial credit is ``flattening'' the target distribution ${\bf q}$ across the set $M_i$: 
\begin{align}
\label{eq:softtarget}
\text{SoftTargetLoss} = H({\bf p}, {\bf q}), \ q_k =  \delta(k \in M_i)/|M_i|
\end{align}
We call this loss {\em soft target}. This setup evenly distributes the probability mass among all the reasonable classes. While boosting the performance at 10,000 classes (Fig \ref{fig:increasingK}), it still fails with larger $K$. We also experimented with Gaussian-weighted soft targets, but found similar trends. We posit that the gradient signal from a flattened target becomes too weak. Over half the training examples contain more than 4,000 memberships. The gradient of the cross-entropy loss is $\frac{\partial H}{\partial s_i(I)} = p_i - q_i$, with $q_i$ becoming dilated, the gradient signal for positive examples vanishes.

\subsection{Attempt 3: Multi-label Targets}

\begin{figure}[t]
\begin{center}
   \includegraphics[width=1\linewidth]{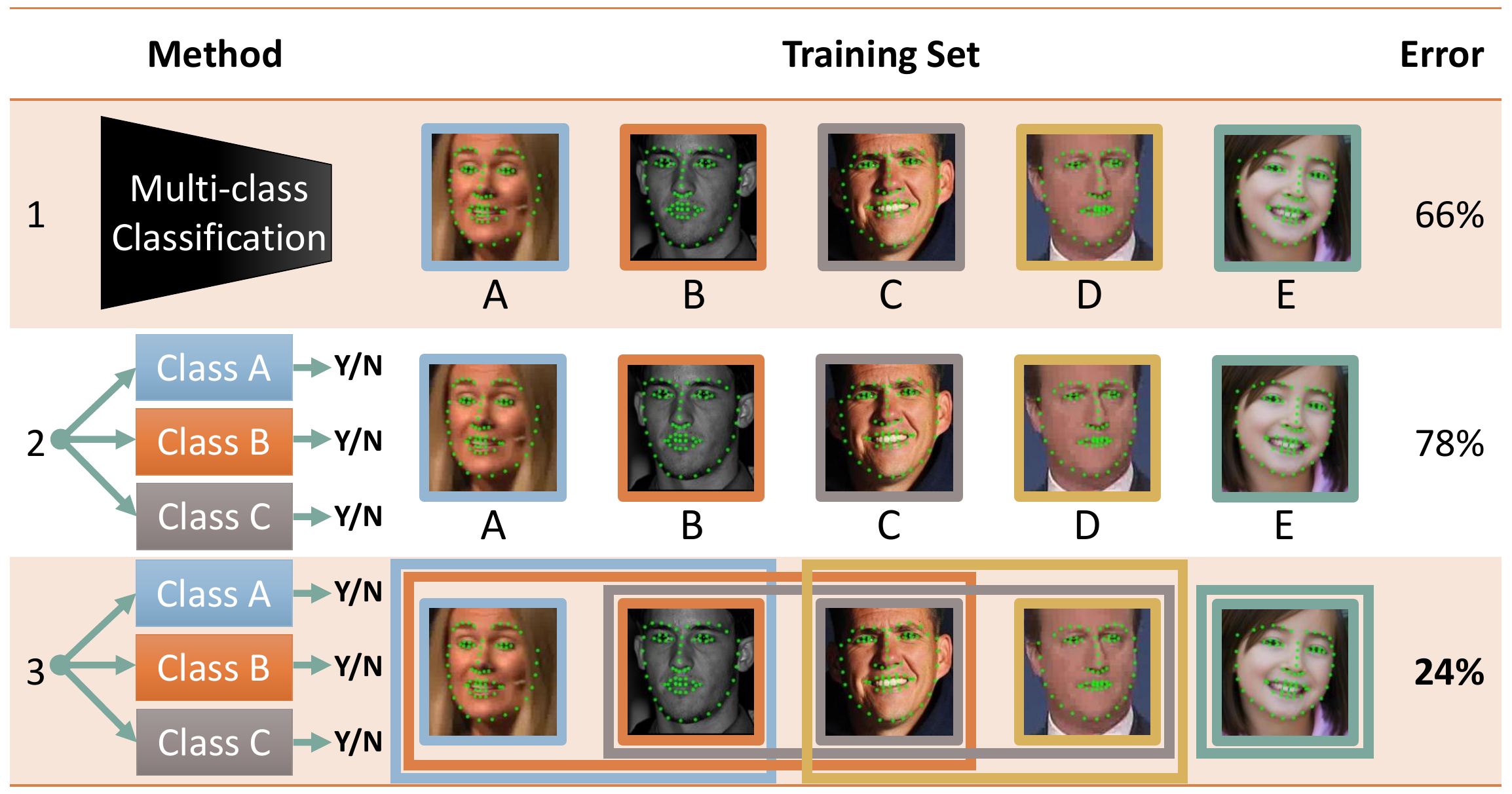}
\end{center}
   \caption{Scaling up the number of classes in a classification network. This figure shows three different ways of training a multi-class classifier, with the mean validation error of the landmarks shown in the last column. The error is shown as a percentage of the error of a random classifier. We adopt the third method for our approach, where we train independent binary classifiers with {\em example sharing}. The colors in the figure denote classes and the boxes in the last row circle the training examples used for a particular class (see the matching color). For example, the blue box denotes that the images of class A and B are used as the {\em positive} examples for training class A.}
\label{fig:classify}
\end{figure}

To allow for training examples with large memberships to still be guided by a strong learning signal, we could treat the $K$ target classes as $K$ separate binary prediction problems, sometimes known as {\em multi-label} classification:
\begin{align}
\label{eq:multilabel}
    \text{MultiLabelLoss} = -\sum_{k = 1}^{K}\log(1 + \exp(-c_k s_k({\bf I}))), \\
    c_k = +1 \text{ if } k \in M_i, -1 \text{ if } k \notin M_i.
\end{align}
Now a training example provides an independent gradient signal to multiple classes at the same time. Importantly, the magnitude of the training signal will not be weakened by the number of neighboring classes. This implies that the $K$-way classification problem (where classes are mutually-exclusive) can be reduced to $K$ independent binary classification problems (where classes can overlap in concept) for training. At test-time, we output a single class label by replacing the proposed loss layer with a softmax layer. 

{\bf Performance:} The performance of the losses are summarized in Fig \ref{fig:increasingK}. The results are striking - multi-label loss continually increases in performance with larger $K$, even at the extreme exemplar class setting. The problem now appears much easier to optimize. Importantly, this performance increase does not come from the loss itself. Fig \ref{fig:classify} compares standard $K$-way to multi-label training {\em without} overlapping pose classes, in which case multi-label learning {\em hurts} accuracy. 
Rather, the particular combination of multi-label learning and overlapping clusters appears to be crucial for learning at scale.

{\bf Analysis:} Our surprising results are consistent with those reported in~\cite{zhu2014capturing}, who demonstrate that adding additional closeby examples to exemplar detectors acts a regularizer that pulls the classifiers towards the class mean (rather than pulling classifier toward the zero-vector, as standard $L_2$ regularization does). When examining Fig \ref{fig:imgnbcnt}, it becomes clear when multi-label learning with overlapping classes, certain positive examples have a dramatically larger impact than others. For example, the frontal-neutral image appears 10,000 times more often than the extreme profile image in the targets. In $K$-way classification, both images have equal impact. The uniform impact holds even for soft targets, since the total influence of an example sums to 1. For those interested in cognitive motivations, our results might be consistent with prototype theories of mental categorization~\cite{rosch1978cognition}, which also suggest that some examples from a category are more prototypical than others (and so perhaps should have a bigger impact during learning).


\begin{table}[t]
\centering
\begin{tabular}{l|lllll} \hline
Method     & Linear & NN (${\bf w}$) & NN ($s$) \\ \hline
Pt-Pt Error & 0.0205 & 0.0206   & 0.0222 \\ \hline
\end{tabular}
\caption{Comparison of our classification network with nearest neighbor classifier. Here nearest neighbor classifiers use cosine distance. The comparable performance shows that the features themselves are high quality summarization of the face images. In the exemplar setup, the classification filter weights can be viewed as an embedding of the training set.}
\label{tab:nn}
\end{table}

{\bf Comparisons to nearest-neigbors:} Now that we have a scalable exemplar-class model, it is interesting to compare it to classic approaches for nearest-neighbor (NN) learning. NN-based learning is often addressed as a metric-learning problem.
Naively, one can consider the features before the classification layer $s({\bf I})$ to be a good embedding of the original image, (in the ResNet-50 case, the pool-5 layer). We find through our experiments that the filter weights ${\bf w}$ of the classification layer might be a better embedding. 
This makes sense in retrospect: the final class is computed by taking the dot product of the filter weights and the feature vector and adding a bias, and then maxing across classes. 
In practice, the bias is usually zero, implying that the exemplar-class score resembles a cosine similarity.  To verify this idea, we extract the features from the validation set and classify them in the nearest neighbor fashion using (a) their classifier weights and (b) their pool-5 features.
Table \ref{tab:nn} suggests that exemplar-classification may be an alternate methods of learning embeddings. 


\subsection{Pre- and Post-Processing}

\begin{figure}[t]
 \begin{center}
   \includegraphics[width=1\linewidth]{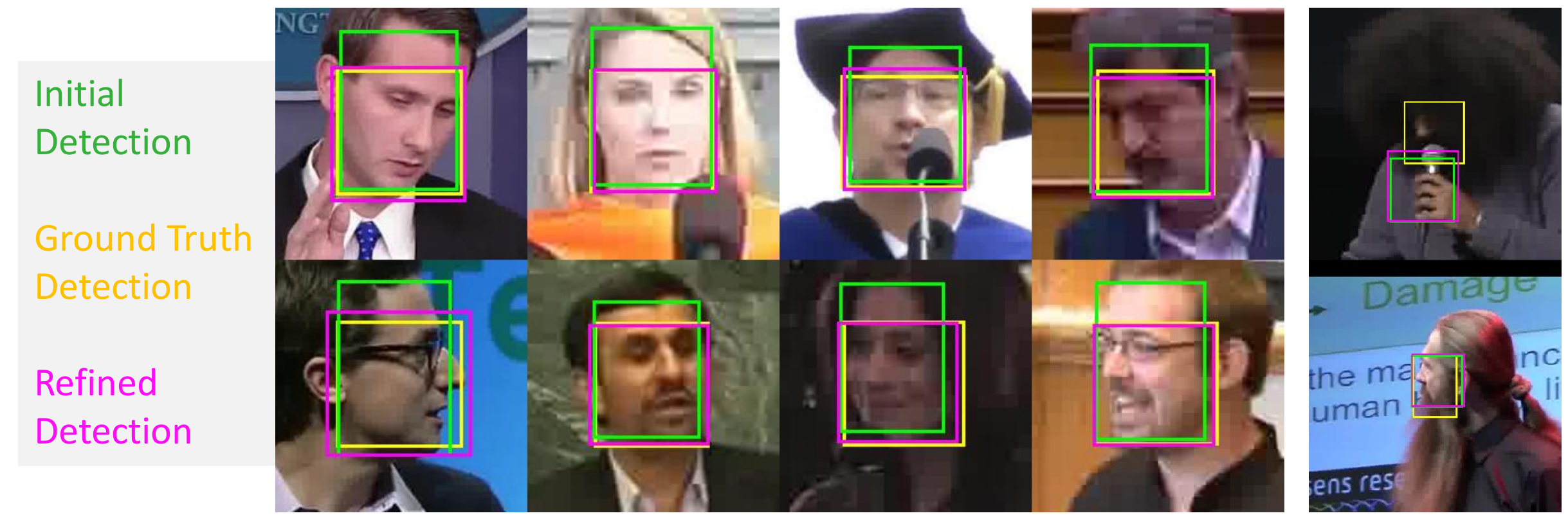}
 \end{center}
   \caption{Detection refinement. We train a linear regressor on top of the features of our exemplar model. Even though our model is trained in a classification setup, spatial information is learned useful for the detection refinement. Failure cases are shown on the rightmost column.}
 \label{fig:detref}
 \end{figure}

{\bf Detection Refinement (pre):}
Similar to other face alignment systems, we assume a rough detection of the face provided. We find that off-the-shelf detectors produce bounding boxes that are off in location and size comparing to the ground truth bounding box. Therefore, we learn a linear bounding-box regressor (similar to R-CNN \cite{girshick14CVPR}) that uses features from our classification network to refine detection windows (Fig \ref{fig:detref}). We then feed the refined detection image region into our classification network.

{\bf Pose Class Regressors (post):}
Though our pose classifier works well (as evidenced by its lowest failure rate on the benchmark), it struggles to produce accurate fine-scale predictions.
To alleviate this, for each pose class $k$, we train a cascaded linear regressor \cite{Xiong2013} using all the member images $\{i: k \in M_i\}$. 
Ideally, we want to train a regressor for each class. In reality, we share the weights among the classes and only train a small number of regressors (100) due to computational constraints. Since we use exemplar-class in our final model, the clustering of the exemplar classes is equivalent to the clustering of training examples as explained in Sec \ref{sec:softtarget}
(\ie, k-means clustering).  Note that the same example sharing takes place when training the regressors.


\begin{figure*}[t]
    \centering
    \begin{subfigure}[b]{0.3\textwidth}
        \centering
        \includegraphics[width=\textwidth]{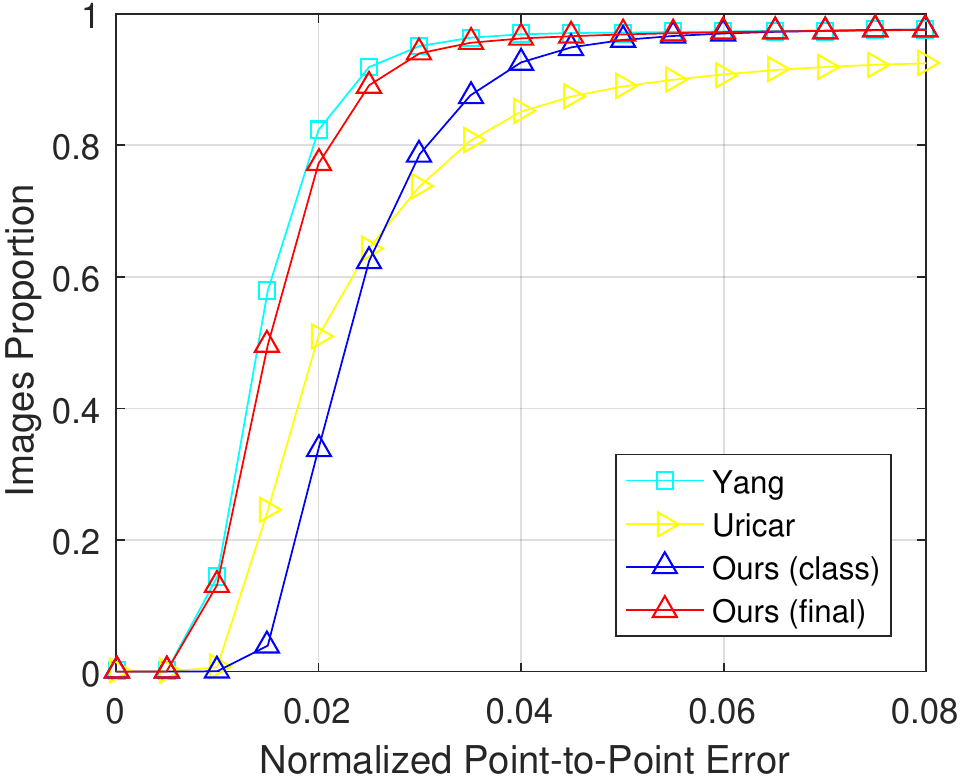}
            \caption{{\small Category 1 - naturalistic and well-lit.}}
        \end{subfigure}
        \qquad
        \begin{subfigure}[b]{0.3\textwidth}   
            \centering 
            \includegraphics[width=\textwidth]{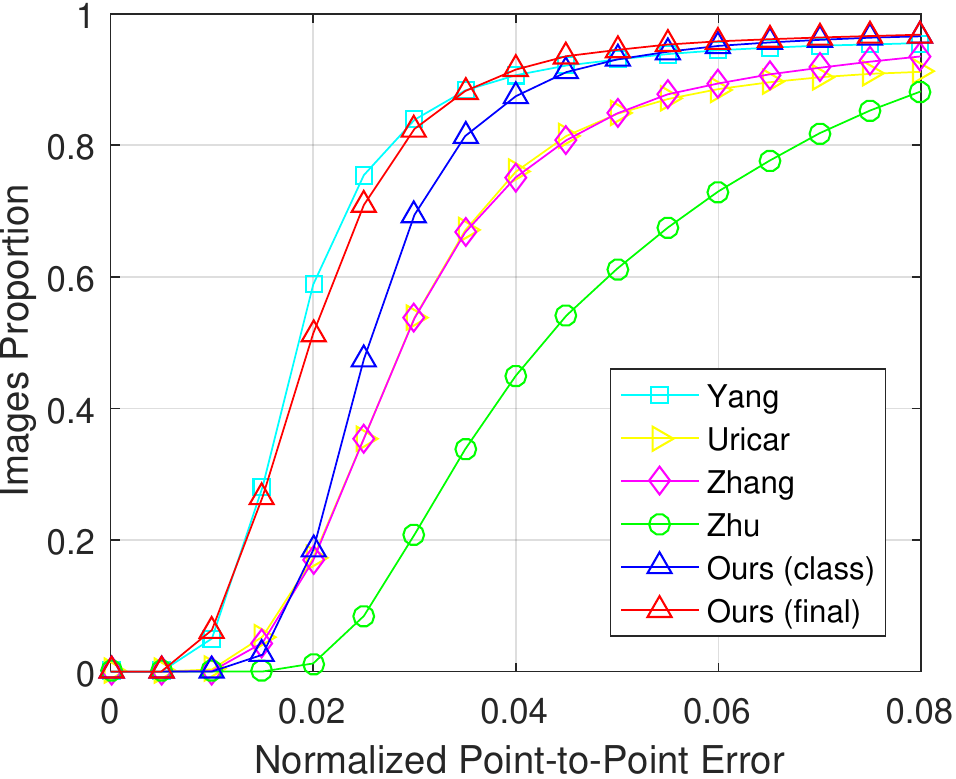}
            \caption{{\small Category 3 - unconstrained}}    
            \label{fig:test4}
        \end{subfigure}
        \qquad
        \begin{subfigure}[b]{0.3\textwidth}   
            \centering
            \includegraphics[width=\textwidth]{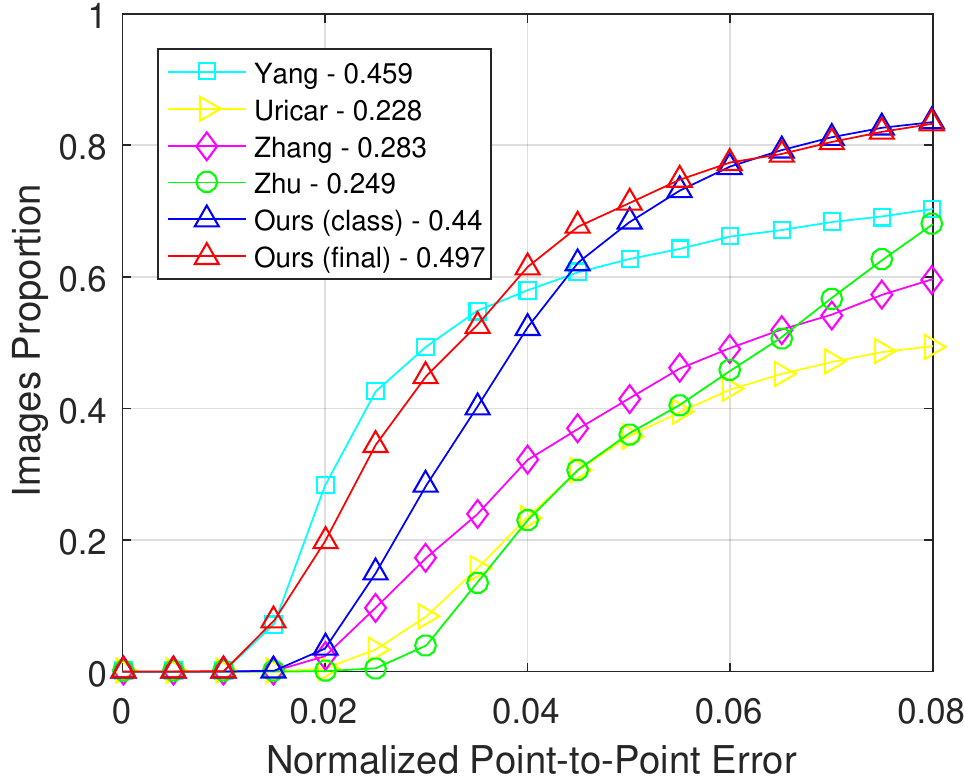}
            \caption{{\small Subset of cat 3 - hard frames only}}    
            \label{fig:test6}
        \end{subfigure}
    \caption{\small The cumulative error distribution curves on the 300VW benchmark. The statistics for (a) and (b) are summarized in Tab \ref{tab:300vw} . The Area Under Curve (AUC) in (c) is show next to the names.}
    \label{fig:curves}
\end{figure*}

%

{\bf Temporal Smoothing (post):} As previously shown, our model is capable of global uncertainty reasoning and we exploit this in combination with temporal smoothing to achieve better results on video datasets. Given the distribution over $K$ poses in consecutive frames, we can easily construct a $K$-state hidden-Markov model (HMM) that only allows transitions between similar pose classes \eqref{eq:members}. We can then use max-product inference to {\em decode} a sequence of temporally-smooth, high-scoring pose classes.

\section{Experiments}
\setcounter{secnumdepth}{2}


    


\subsection{Standard Benchmark Evaluation}

{\bf Datasets:}
We test our algorithms on the 300VW dataset \cite{shen2015first}, a standard benchmark for video face alignment. It contains 114 videos, 50 of which are used for testing. The test set is partitioned into three categories with varying degree of occlusion, pose variation, and extreme illumination. The category 1 is considered the most constrained while the category 3 the most unconstrained. Since the performance of category 1 and 2 are similar, we report the performance of category 1 and 3 here and refer readers to the supplement for the complete evaluation. Furthermore, in the next section, we compare our method with existing algorithms on the hardest frames in category 3 (most challenging) as an auxiliary benchmark.

To form the validation set, we randomly pick 10\% of the training videos. For the remaining 59 training videos, we subsample 10\% of the frame at uniform interval to remove data correlation. This forms our base training set. As is {\em standard practice}, we include into our training set images from 300W \cite{Sagonas2013300FI}, IBUG, HELEN, LFPW, AFW \footnote{Additional datasets were allowed in the original challenge.}. Moreover, we include synthetic large pose dataset 300W-LP \cite{Zhu2016FaceAA}. Note this dataset is in fact an augmentation from the real datasets listed above, and therefore, {\em we include no extra supervision compared to the standard practice}. Our final training set has the composition of 8,389 video frames, 4,437 real images, and 61,225 synthetic images. With left-right flip augmentation, we arrive at 140,428 training images in total.

\definecolor{cfirst}{rgb}{0.9, 0.3, 0.3}
\definecolor{csecond}{rgb}{1.0000    0.8314    0.1647}
\definecolor{cthird}{rgb}{0.3333    0.6667    1.0000}
\newcommand{\fst}{\cellcolor{cfirst}}
\newcommand{\snd}{\cellcolor{csecond}}
\newcommand{\trd}{\cellcolor{cthird}}

\begin{table}[t]
\centering
\resizebox{.48\textwidth}{!}{
\begin{tabular}{l||r|r|r|r}
\hline
Method &
C1 AUC & 
C1 FR &
C3 AUC &
C3 FR \\
\hline
\cite{Chrysos2017ACP}
& 0.748     & 6.055     & \fst 0.726  & 4.388 \\ \hline
\cite{Yang2015FacialST}
& \fst 0.791  & \trd 2.400     & \trd 0.710     & 4.461 \\ \hline
\cite{Uricr2015FacialLT}
& 0.657     & 7.622     & 0.574     & 7.957 \\ \hline
\cite{Xiao2015FacialLD}
& 0.760     & 5.899     & 0.695     & 7.379 \\ \hline
\cite{Rajamanoharan2015MultiviewCL}
& 0.735     & 6.557     & 0.659     & 8.289 \\ \hline
\cite{Wu2015ShapeAR}
& 0.674     & 13.925    & 0.602     & 13.161 \\ \hline
\cite{Zhang2014FacialLD}
& N/A       & N/A       & 0.409    & 6.487 \\ \hline
\cite{Zhu2016FaceAA}
& N/A       & N/A       & 0.635    & 11.796 \\ \hline
Ours (classification)
& 0.678     & \snd 2.398     & 0.635    & \trd 3.431 \\ \hline
Ours (+ regressor)
& \trd 0.774     & \fst 2.221  & 0.709    & \fst 3.189 \\ \hline
Ours (+ temp. smooth.)
& \snd 0.777     & 2.462    & \snd 0.718     & \snd 3.298   \\ \hline
\end{tabular}
}
\caption{Comparing with existing methods on the 1st and 3rd category of 300VW benchmark. The \textcolor{cfirst}{1st}, \textcolor{csecond}{2nd} and the \textcolor{cthird}{3rd} place for each metric are color coded. Here AUC denotes the area under the CED curves in Fig \ref{fig:curves} and FR denotes the failure rate in percentage.}
\label{tab:300vw}
\end{table}

{\bf Implementation details:}
For this experiment, we use the exemplar-class, since we find the more classes, the lower error the model will predict.
The membership set threshold $\tau$ is determined through validation. When training the exemplar classifier, we use the ground truth detection. At test time, we run an out-of-the-shelf detector \cite{Zhang2016SP} before applying our detection refinement. The detector is trained on the CelebA \cite{liu2015deep} and the WIDER FACE datasets \cite{yang2016wider}. We evaluate our model in a detection setup as opposed to a tracking setup because the detection setup is simpler and it is reported that tracking with failure detection provides only marginal improvement over the detection setup \cite{Chrysos2017ACP}. For the post-processing fine-tuning, we train 100 pose class regressors with 7 levels of cascades. For temporal smoothing, we use a low pass filter on 3 consecutive frames. Our MATLAB code takes around 60ms per frame, including detection refinement and post-processing regressors.

\begin{figure}[t]
\begin{center}
   \includegraphics[width=1\linewidth]{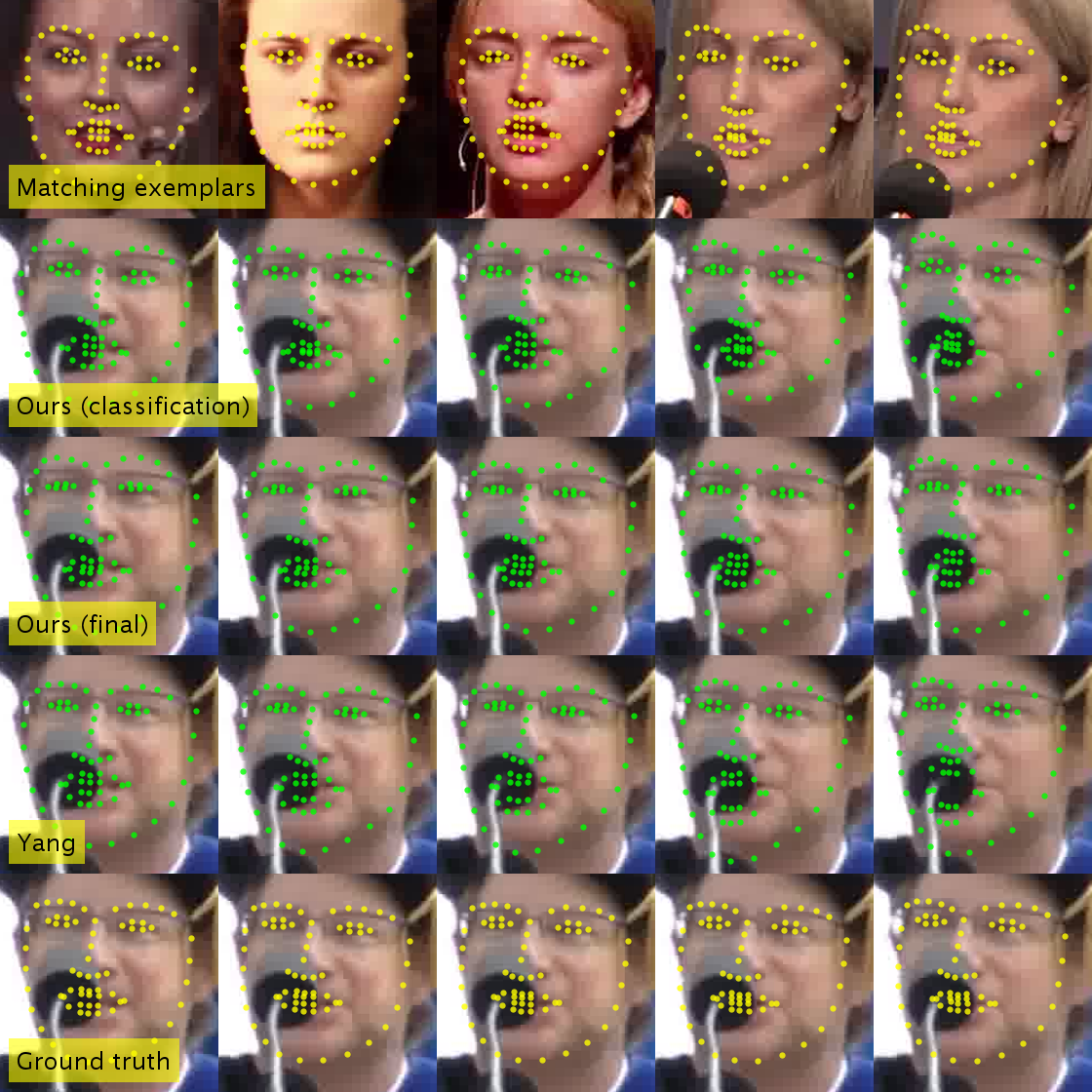}
\end{center}
   \caption{Qualitative results for Category 3 frames on 300VW. The competing method \cite{Yang2015FacialST} fails when the occluder moves towards the center of the face while our methods can still detect the landmarks. While our classification produces reasonable prediction (Row 2), the post-processing regressors and temporal smoothing technique (Row 3) fix small localization error. We also visualize the training exemplar associated with each classification.}
\label{fig:300vw_vis}
\end{figure}

{\bf Comparing to the state-of-the-art:}
We follow the updated standard of the 300VW benchmark \cite{Chrysos2017ACP}, where all frames are included in the evaluation and the metric is point-to-point error (or also RMSE) normalized by the diagonal of the ground truth bounding box. The Cumulative Error Distribution (CED) curves are provided in Fig \ref{fig:curves} (a, b), while the Area Under Curves (AUCs) and failure rates are summarized in Tab \ref{tab:300vw}\footnote{The evaluation is done in the standard 68-pt format. We compared with another state-of-the-art method, iCCR \cite{SnchezLozano2016CascadedCR}, in their 66-pt format. Our method consistently outperforms iCCR. Details can be found in the supplement.}. We focus on the {\bf failure rate}, which the benchmark defines to be the fraction of images with normalized error above 0.08.
Methods in the original 300VW challenge are evaluted by the challenge organizers. We include two additional methods \cite{Zhu2016FaceAA,Zhang2014FacialLD} for which we find the testing code available.

The standard evaluation shows our method reaches the state-of-the-art performance consistently in constrained and unconstrained setup. Moreover, our method achieves much lower failure rate on the most challenging category, indicating the robustness of approach. This is only made possible through large scale classification. For applications that require fine-scale landmark localization, we adopt regression and temporal smoothing to fine-tune the landmarks, while still maintaining the exceptional low failure rate. Visual examples are shown in Fig \ref{fig:300vw_vis}.

\begin{figure}[t]
\begin{center}
   \includegraphics[width=1\linewidth]{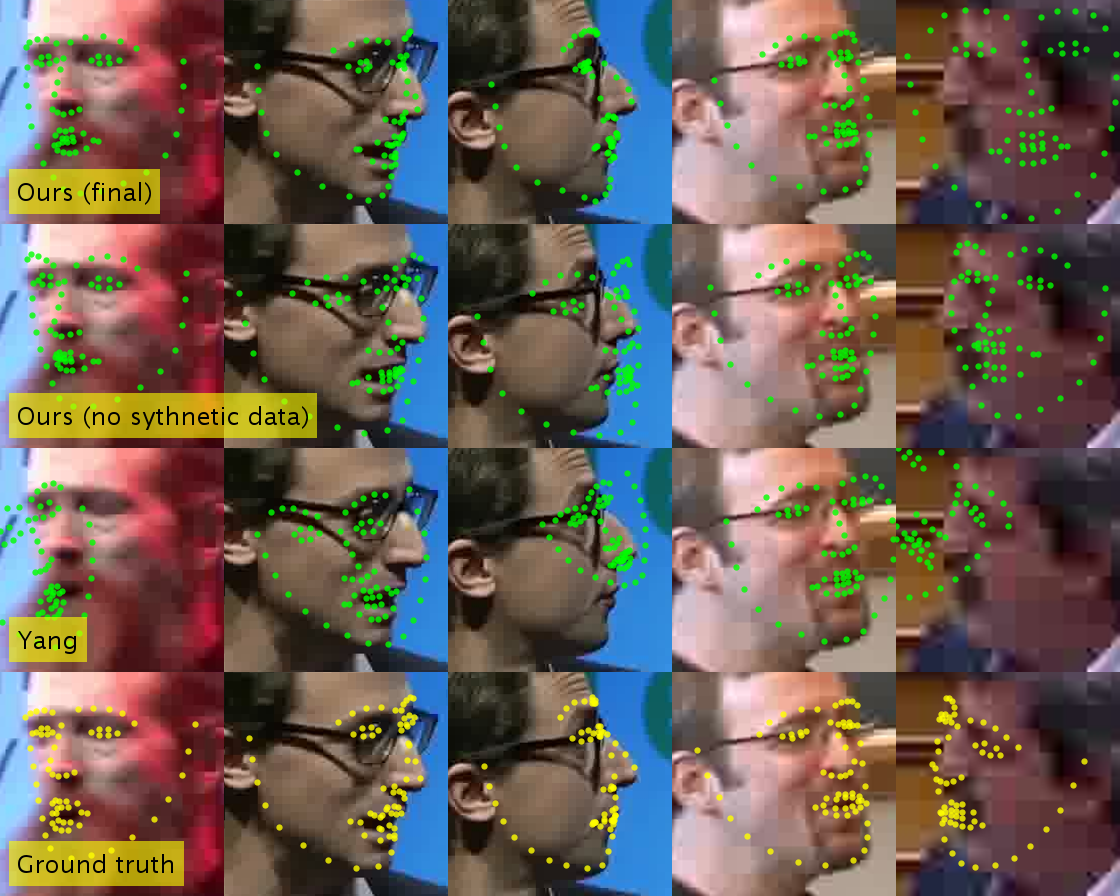}
\end{center}
   \caption{Qualitative results for Category 3-Hard frames on 300VW. We can see that by training on synthetic images (comparing Row 1 with Row 2), our method is robust to large pose variation. The last column shows a failure case.}
\label{fig:300vw_vis_hard}
\end{figure}


\subsection{Hard Case Evaluation}
Because many algorithms do well on most of the frames, prior work has evaluated on subsets of challenging frames with larger yaw variations~\cite{Zhu2016FaceAA}. We include this evaluation in supp material, for which we dramatically outperform past work. Inspired by this, we systematically construct the 10\% of the frames from category 3 that deviate the most from the average shape (and so also include variations in pitch, expression, \etc). Results on this hard subset are presented in Fig \ref{fig:curves} (c) and Fig \ref{fig:300vw_vis_hard}. {\em Our approach has a considerably lower error rate (and higher AUC) than all prior work on this difficult subset}, indicating the robustness of a classification approach. Such robustness will prove crucial for many applications that process ``in-the-wild'' data, such as gaze prediction for understanding social interaction.

\begin{figure}[b]
\begin{center}
   \includegraphics[width=1\linewidth]{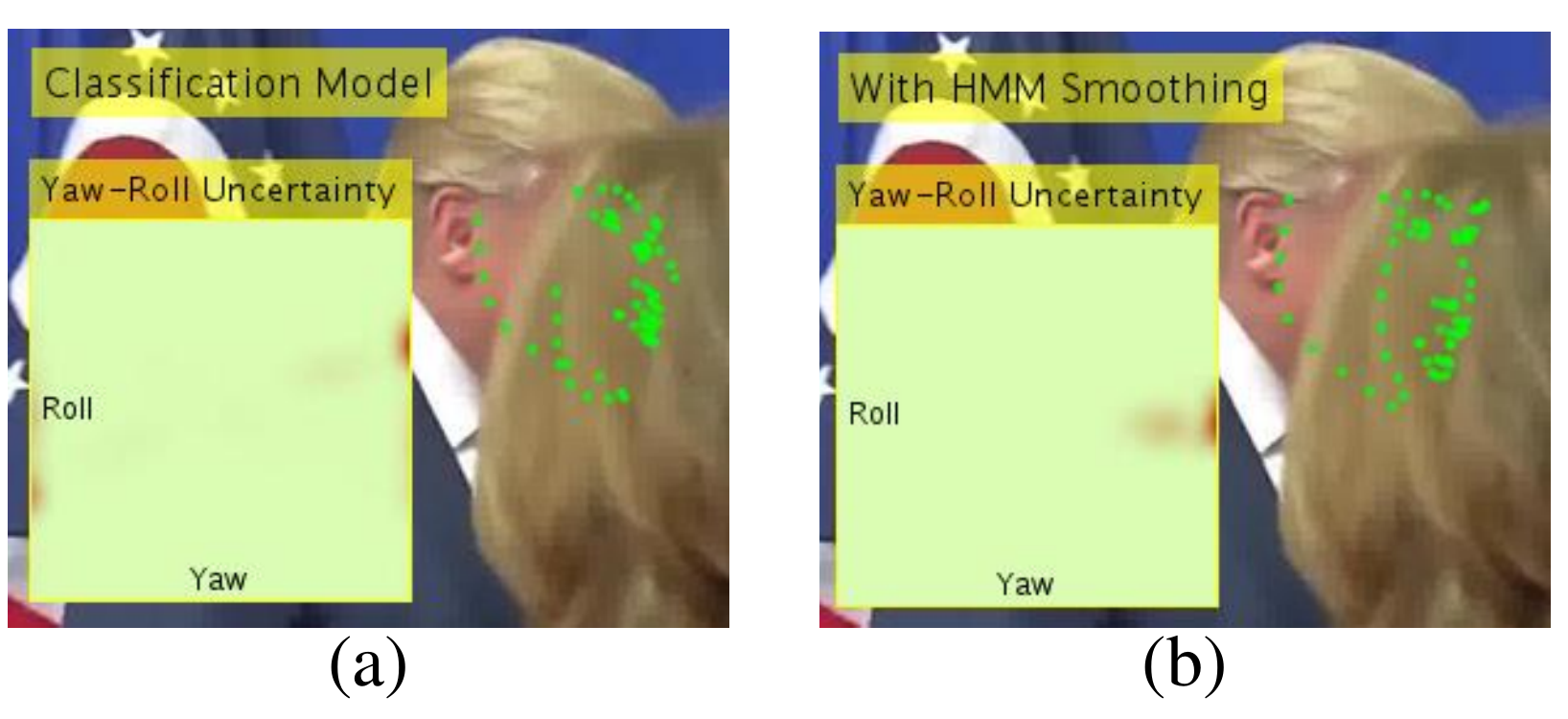}
\end{center}
   \caption{Uncertainty reasoning under occlusion. Our classification model ({a}) can report uncertainty in terms of global variables, \eg yaw and roll. By integrating temporal information over time with an HMM ({b}), we can further increase accuracy and reduce uncertainty during such occlusions.}
\label{fig:compocclusion}
\end{figure}

To further illustrate the robustness and the uniqueness of our model, we evaluate results on a challenging web video with clutter and occlusion (Fig \ref{fig:compocclusion}). By decoding the classification output with temporal information, our approach can recover the rough pose even under complete occlusion. Such decoding is made possible only by our uncertainty representation, which is not found in existing methods. 
The results on the entire video can be found on the author's website.

{\bf Is the problem of face alignment solved?} It is widely considered that problem of face alignment with small pose and perfect lighting is solved. However, it might not be the case when the scenario becomes more complicated. We find that there are many frames in the hard subset where all methods fail, meaning they cannot estimate even a rough pose (one example shown in the last column of Fig \ref{fig:300vw_vis_hard}). These frames usually consist of large pose variation in combination with low resolution, extreme lighting and motion blurs. These factors pose a challenge not only to face alignment, but also face detection and tracking algorithms. 

\subsection{Interactive Annotation}
\label{sec:interactive}
\begin{figure}[t]
\begin{center}
   \includegraphics[width=1\linewidth]{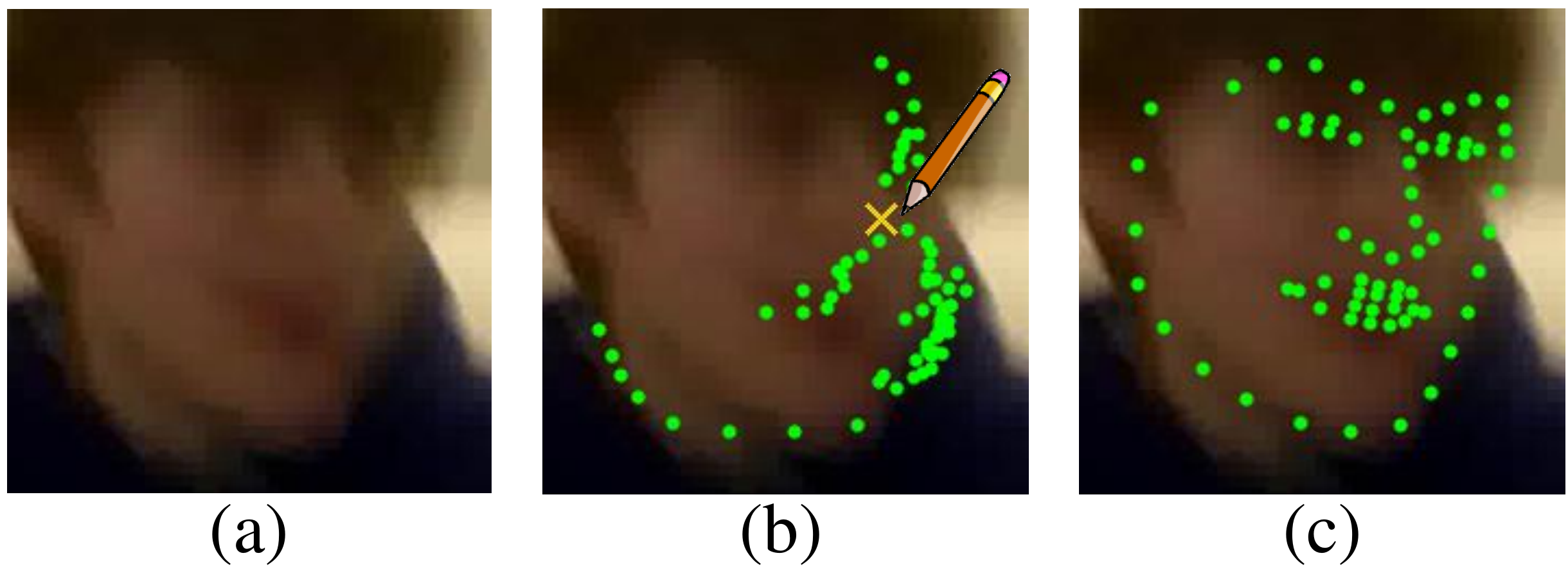}
\end{center}
   \caption{Conditional prediction. Our algorithm fails on images with severe motion blur (a) \& (b).  However, if an annotator gives hint of where the nose tip is (the yellow cross in (b)), we can correct the mistake (c) by finding the most probable class {\em conditioned} on this evidence.}
\label{fig:condpred}
\end{figure}

Recent work has suggested that interactive annotation can significantly improve the efficiency of labeling new datasets~\cite{le2012interactive}. Our model can be used for this purpose through conditional prediction (Fig~\ref{fig:condpred}). Given evidence $E$ provided by a user (\eg, a single landmark click), compute the subset of pose-classes consistent with that evidence $\Omega(E)$, and return the normalized softmax distribution over this subset:
\begin{align}
   p({\bf y}|E) \propto \sum_{k \in \Omega(E)}  p_k \phi_k(||{\bf y}-{\bm \mu}_k||).
\end{align}
In contrast, it is unclear how to incorporate such interactive evidence into regression-based methods such as CRMs.

We conduct an experiment to evaluate the impact of interactive annotation in Tab \ref{tab:condpred}. In summary, by simply asking a user to annotate a single landmark (the nose) in each frame, one can reduce the error rate by 2-fold and correctly annotate 98.5\% of the frames. By selecting an {\em optimal} landmark to label (through active learning on top of our probabilistic outputs), one can potentially reduce error by another factor of 2. Two demo videos can be found on the author's website showing our interactive annotation in progress.
Finally, when annotating a video dataset, similar approaches can be used to actively select both the key frame and key point that will be most informative (when combined with a HMM to obtain predictions for all other frames).


\begin{table}[t]
\centering
\begin{tabular}{l||r}
\hline
Method &
Failure rate (\%) \\
\hline
No annotation (baseline)
& 4.010 \\ \hline
1-pt conditioning
& 1.564 \\ \hline
Best 1-pt (upper bound)
& 0.639 \\ \hline
\end{tabular}
\caption{We simulate interactive annotation of all category 3 frames in 300VW. ``1-pt'' refers to a user labeling a fixed landmark (the nose) in each frame, while ``best'' refers to an upper-bound obtained labeling the optimal landmark minimizing the error. Our results suggest that with a single user-click per image (as opposed to 68 landmark clicks), one can correctly label 99.4\% of the frames.}
\label{tab:condpred}
\end{table}

\section{Conclusion}
\setcounter{secnumdepth}{2}
Though visual alignment is naturally cast as a regression problem, we reformulate it as a classification task. One significant advantage is that softmax classification networks naturally report back distributions, which can be used to reason about confidence, uncertainty, and condition on external evidence (provided by contextual constraints or an interactive user). Despite its simplicity, such a method is considerably more robust than prior work, producing state-of-the-art accuracy in challenging scenarios. We focus on the illustrative task of facial landmark alignment, demonstrating robust performance across large pose variation, severe occlusions and extreme illumination. 


{\bf Acknowledgements:}
This research was supported in part by the National Science Foundation (NSF) under grant IIS-1618903, the Defense Advanced Research Projects Agency (DARPA) under Contract No. HR001117C0051, and the Office of the Director of National Intelligence (ODNI), Intelligence Advanced Research Projects Activity (IARPA), via IARPA R \& D Contract No. D17PC00345. Additional support was provided by Google Research and the Intel Science and Technology Center for Visual Cloud Systems (ISTC-VCS). The views and conclusions contained herein are those of the authors and should not be interpreted as necessarily representing the official policies or endorsements, either expressed or implied, of ODNI, IARPA, or the U.S. Government. The U.S. Government is authorized to reproduce and distribute reprints for Governmental purposes notwithstanding any copyright annotation thereon.

\bibliography{egbib}
\bibliographystyle{aaai}

\end{document}